\documentclass[conference]{IEEEtran}
\usepackage{times}
\usepackage[numbers]{natbib}
\usepackage{multicol}
\usepackage[bookmarks=true]{hyperref}

\usepackage[utf8]{inputenc}
\usepackage{graphicx}
\usepackage{multicol}
\usepackage{footmisc}
\usepackage{amssymb}
\usepackage{amsmath}
\usepackage{amsfonts}
\usepackage[ruled]{algorithm}
\usepackage{bbm}
\usepackage{tablefootnote}
\usepackage{multirow}
\usepackage[noend]{algorithmic}
\usepackage[dvipsnames]{xcolor} 
\usepackage{tikz}
\usepackage{float}
\usepackage{hyperref}
\usepackage{booktabs} 

\usetikzlibrary{math,shapes,positioning}
\usepackage[font={small}]{caption}

\usepackage{subcaption}
\usepackage{listings}

\newcommand{\clipart}[1]{\includegraphics[height=6mm]{images/#1.png}}

\usepackage{color, soul} 
\renewcommand\hl[1]{#1} 
\IEEEoverridecommandlockouts
\begin{document}


\title{FlowBot3D: Learning 3D Articulation Flow to Manipulate Articulated Objects}

\author{Ben Eisner$^*$\thanks{*Equal contribution.}, Harry Zhang$^*$, David Held\\
Carnegie Mellon University\\
Pittsburgh, PA, USA\\
\texttt{\{baeisner, haolunz, dheld\}@andrew.cmu.edu}
}



%

\maketitle

\begin{abstract}
We explore a novel method to perceive and manipulate 3D articulated objects that generalizes to enable a robot to articulate unseen classes of objects. We propose a vision-based system that learns to predict the potential motions of the parts of a variety of articulated objects to guide downstream motion planning  of the system to  articulate the objects. To predict the object motions, we train a neural network to output a dense vector field representing the point-wise motion direction of the points in the point cloud under articulation. We then deploy an analytical motion planner based on this vector field to achieve a policy that yields maximum articulation. We train a single vision model entirely in simulation across all categories of objects, and we demonstrate the capability of our system to generalize to  unseen object instances and novel categories in both simulation and the real world using the trained model for all categories, deploying our policy on a Sawyer robot with no finetuning. Results show that our system achieves state-of-the-art performance in both simulated and real-world experiments. Code, data, and supplementary materials are available \href{https://sites.google.com/view/articulated-flowbot-3d/home}{\textcolor{blue}{here}}.
\end{abstract}

\IEEEpeerreviewmaketitle

\section{Introduction}
\label{sec:introduction}

Understanding and being able to manipulate articulated objects such as doors and drawers is a key skill for robots operating in human environments. While humans can rapidly adapt to novel articulated objects, constructing robotic manipulation agents that can generalize in the same way poses significant challenges,
since the complex structure of such objects requires three-dimensional reasoning of their parts and functionality. Due to the large number of categories of such objects and intra-class variations of the objects' structure and kinematics, it is difficult to train efficient perception and manipulation systems that can generalize to those variations.


To address these challenges, we propose to separate this problem into one of ``affordance learning" and ``motion planning."
If a robot can predict the potential movements of an objects' parts (a.k.a. ``affordances"), it would be relatively easy for the agent to derive a downstream manipulation policy by following the predicted motion direction. Thus, we tackle the problem of manipulating articulated objects by learning to predict the motion of individual parts on articulated objects.


\begin{figure}
    \centering
    \includegraphics[width=\linewidth]{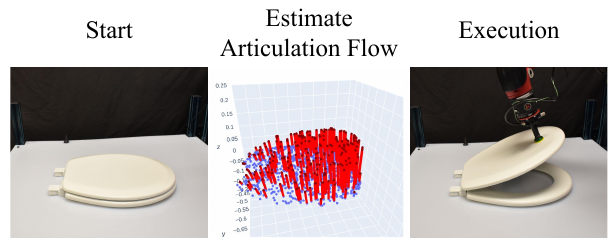}
    \caption{FlowBot3D in action. The system first observes the initial configuration of the object of interest, estimates the per-point articulation flow of the point cloud (3DAF), then executes the action based on the selected flow vector. Here, the red vectors represent the direction of flow of each point (object points appear in blue); the magnitude of the vector corresponds to the relative magnitude of the motion that point experiences as the object articulates. }
    \label{fig:teaser}
\end{figure}


Previous work has proposed to learn the articulation parameters (i.e. rotation axis of revolute joints and translation axis of prismatic joints) in order to guide the manipulation policy \cite{Cheong2007-iw}. However, such methods often rely on knowing class-specific articulation structures. Without such knowledge, the policies can neither operate nor be applied to novel categories.

To learn a generalizable perception and manipulation pipeline, we need to be robust to the variations of the articulated objects' geometries and kinematic structures. We seek to construct a vision system that can learn to predict how the parts move under kinematic constraints without explicitly knowing the articulation parameters: specifically, the location of the rotational or translational axes for revolute or prismatic parts, respectively.

In this paper, we present FlowBot3D, a deep 3D vision-based robotic system that predicts dense per-point motion of an articulated object in 3D space, and leverages this prediction to produce actions that articulate the object. We define such per-point motion as the \textbf{3D articulation flow (3DAF)} vectors, since this representation describes how each observed point on the articulated part would ``flow'' in the 3D space under articulation motion. Such a dense vector field prediction can then be used to aid downstream manipulation tasks for both grasp point selection as well as predicting the desired robot motion after grasping.
We train a \textit{\textbf{single}} 3D perception module to perform this task across many object categories, and show that the trained model generalizes to a wide variety of objects -- both in seen categories, and entirely unseen object categories.

The contributions of this paper include:
\begin{enumerate}
    \item A novel per-point representation of the articulation structure of an object, 3D Articulation Flow (3DAF).
    \item A novel 3D vision neural network architecture (which we call ArtFlowNet) that takes as input a static 3D point cloud and predicts the 3D Articulation Flow of the input point cloud under articulation motion.
    \item A novel robot manipulation system (FlowBot3D) for using the predicted 3D Articulation Flow to manipulate articulated objects.
    \item Simulated experiments to test the performance of our system in articulating a wide range of PartNet-Mobility dataset objects.
    \item Real-world experiments deployed on a Sawyer robot to test the generalizablity and feasibility of our system in real-world scenarios. 
\end{enumerate}

\section{Related Work}
\label{sec:related}

\textbf{Articulated Object Manipulation}: Manipulation of articulated objects and other objects with non-rigid properties remains an open research area due to the objects' complex geometries and kinematics. Previous work proposed manipulating such objects by hand-designed analytical methods, such as the immobilization of a chain of hinged objects by \cite{Cheong2007-iw, zhang2016health, zhang2021robots}.
 \cite{berenson2011task, zhang2020dex, lim2021planar, lim2022real2sim2real} proposed a planning framework for manipulation under kinematic constraints. \citet{Katz2008-jo} proposed a method to learn such manipulation policies in the real-world using a grounded relational representation learned through interaction. 
 
 With the development of larger-scale datasets of articulated objects such as the PartNet dataset by \citet{Mo2019-az} and Partnet-Mobility by \citet{Xiang2020-oz}, several works have proposed learning methods based on large-scale simulation and supervised visual learning. \citet{Mo2021-jm} proposed to learn articulation manipulation policies through large-scale simulation and visual affordance learning. \citet{Xu2021-iw} proposed a system that learns articulation affordances as well as an action scoring module, which can be used to articulate objects.
 \citet{Mu2021-ui} provided a variety of baselines for the manipulation tasks of 4 categories of articulated objects in simulation.
 Several works have focused specifically on visual recognition and estimation of articulation parameters, learning to predict the pose \cite{Yi2018-mw, Yan2020-hm, Wang2019-gy, Hu2017-bn, Li2020-go, elmquist2022art, eisner2022flowbot3d, avigal20206, avigal2021avplug} and identify articulation parameters \cite{Jain2021-rg, Zeng2020-tk, sim2019personalization} to obtain action trajectories. Moreover, \cite{Narayanan2015-mp, Burget2013-nb, Chitta2010-vn, devgon2020orienting, pan2023tax, zhang2023flowbot++} tackle the problem using statistical motion planning. 


\textbf{Optical Flow for Policy Learning}: Optical flows \cite{Horn1981-ql} are used to estimate per-pixel correspondences between two images for object tracking and motion prediction and estimation. Current state-of-the-art methods for optical flow estimation leverage convolutional neural networks \cite{Dosovitskiy2015-su, Ilg2017-sy, Teed2020-cw, shen2024diffclip, jin2024multi, yao2023apla}. \citet{Dong2021-xe, Amiranashvili2018-vm} use optical flow as an input representation to capture object motion for downstream manipulation tasks. \citet{Weng2021-re} uses flow to learn a policy for fabric manipulation. While the aforementioned optical flows are useful for robotic tasks, we would like to generalize the idea of optical flow beyond pixel space into full three-dimensional space. Instead, we introduce ``3D Articulation Flow'', which describes per-point correspondence between two point clouds of the same object. Another work that is highly related to ours is \citet{Pillai2015-wu}, which learns to predict the articulated objects' parts motion using a motion manifold learner. First, while we both predict the parts' motion to derive an implicit policy, we do not rely on the intermediate articulation parameters in order to predict the motion manifold. Second, we do not rely on any demonstration to learn from - our method learns in a completely self-supervised fashion.

\section{Method - From Theory to Practice}
\label{sec:method}

In this section, we examine the physical task of manipulating the articulation of an articulated object. We first present the theoretical grounding behind the intuition of our method, and we slowly relax assumptions and approximations to create a system that articulates objects in the real world based on point cloud observations. 

\subsection{An Idealized Policy Based On Dynamics and Kinematics}
\label{subsec:3a}

The articulated objects we consider in this work are generally objects that 1) consist of one or more rigid-bodies -- or ``links'' -- which are 2) connected to one another by revolute or prismatic joints with exactly 1 degree of freedom each, and 3) have at least one link rigidly attached to an immovable world frame so that the only motion the object experiences is due to articulation. Each joint connects a \emph{parent link} (often the fixed-world link) and a \emph{child link}, which can move freely subject to the articulation constraints. While these conditions may seem restrictive, under normal ``everyday'' forces many real-world articulated objects (ovens, boxes, drawers, etc.) meet these conditions to a very good approximation.\footnote{We therefore exclude objects with socket joints, free-body objects, and deformable objects from our analysis.} 

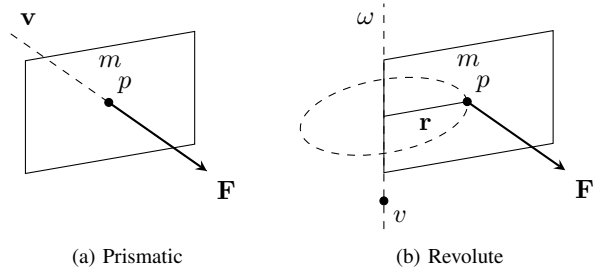
\begin{figure}[h]

    \begin{subfigure}[b]{0.47\linewidth}
         \centering
         
         \begin{tikzpicture}[scale=0.75]
            \draw (0, 1) -- (3, 1.52) node[below, midway]{$m$} -- (3,-0.48) -- (0, -1) -- cycle;
            \filldraw [black, rotate=10] (1.5,0) circle (2pt) node[above right]{$p$}; 

            \draw[-stealth, thick, rotate=10] (1.5, 0) -- (3,-1.5) node[below right]{$\mathbf{F}$};
            \draw[dashed, rotate=10] (0, 1.5) node[above right]{$\mathbf{v}$} -- (3,-1.5) ;
            
            \draw[opacity=0,dashed] (0, -2) -- (0,2);

        \end{tikzpicture}

         \caption{Prismatic}
     \end{subfigure}
     \begin{subfigure}[b]{0.47\linewidth}
         \centering
         
         \begin{tikzpicture}[scale=0.75]
            \draw (0, 1) -- (3, 1.52) node[below, midway]{$m$} -- (3,-0.48) -- (0, -1) -- cycle;
            \filldraw [black, rotate=10] (1.5,0) circle (2pt) node[above right]{$p$}; 
            \filldraw [black] (0,-1.5) circle (2pt) node[below right]{$v$}; 
            \draw[dashed] (0, -2) -- (0,2) node[below left]{$\omega$};
            \draw[-stealth, thick, rotate=10] (1.5, 0) -- (3,-1.5) node[below right]{$\mathbf{F}$};
            
            \draw[rotate=10] (0,0) -- (1.5, 0) node[midway, below]{$\mathbf{r}$};
            \draw[dashed,rotate=10] (0,0) ellipse (1.5 and 0.65);
            
        \end{tikzpicture}

         \caption{Revolute}
     \end{subfigure}
    \caption{Illustrations of prismatic and revolute joints.}
    \vspace*{-10pt}
    \label{fig:idealized}
\end{figure}

We now consider an idealized policy to actuate an articulated object. Suppose we are able to attach a gripper to any point ${p} \in \mathcal{P}$ on the surface $\mathcal{P} \subset \mathbb{R}^3$ of a child link with mass $\mathrm{m}$.  At this point, the policy can apply a 3D force $\Vec{F}$, with constant magnitude $||\Vec{F}|| = C$ to the object at that point.  Our objective is to choose a contact point and force direction  $(p^*, \Vec{F}^*)$ that maximizes the acceleration $\Vec{a}$ of the articulation's child link. If we limit our analysis to two special classes of articulation, revolute joints and prismatic joints, we can very intuitively arrive at the following optimal settings of $(p^*, \Vec{F}^*)$:

\textbf{Prismatic}: A prismatic joint (such as a drawer) can be described as a single 3D unit  vector $\Vec{v}$ which is parallel to its direction of motion. Since motion of the joint is constrained to $\Vec{v}$, the object will provide a responding force $\Vec{F}_n$ to any component of $\Vec{F}$ not parallel to $\Vec{v}$. The net force exerted on the joint by the robot is thus $\Vec{F}_{\text{net}}$, the component of $\Vec{F}$ in the $\Vec{v}$ direction:
\begin{align}
\label{eq:prism}
\Vec{F}_{\text{net}} &= \Vec{F} - \Vec{F}_n \nonumber \\
&= \Vec{F} - (\Vec{F} - (\Vec{F}\cdot\Vec{v})\Vec{v}) = (\Vec{F}\cdot \Vec{v})\Vec{v} = \mathrm{m}\Vec{a}   
\end{align}

As one might expect, the force vector $\Vec{F}^*$ which maximizes the acceleration $\Vec{a}$ occurs when $||\Vec{F}^* \cdot \Vec{v}|| = C$,  i.e. when $\Vec{F}^*$ is parallel to $\Vec{v}$. Because each point $p \in \mathcal{P}$ moves in parallel, applying the force at any point $p$ on the surface will yield the maximum acceleration. Thus, the optimal policy to articulate a prismatic joint is to select any point on the surface and apply a force parallel to $\Vec{v}$ at every time step.

\textbf{Revolute}: A revolute joint (such as a door hinge) can be parameterized by a pair $(v, \boldsymbol\omega)$, where $\boldsymbol\omega$ is a unit vector representing the direction of the axis of rotation about which the child link moves, and $v\in\mathbb{R}^3$ is a point in 3D space that the axis of rotation passes through. Each point $p$ on the child link is constrained to move on the 2D circle perpendicular to the axis of rotation with radius $\Vec{r}$ (where $||\Vec{r}||$ is the length of the shortest vector from $p$ to the line given by $f(t) = v + t\boldsymbol\omega$). Given any point $p$, we can maximize the acceleration by a similar argument as before, except any force in the direction of $\Vec{r}$ or $\boldsymbol\omega$ will be resisted:
\begin{align}
\label{eq:rot}
\Vec{F}_{\text{net}} = \Vec{F} - \Vec{F}_n = \Vec{F} - \left(\frac{\Vec{F} \cdot \Vec{r}}{||\Vec{r}||^2}\right)\Vec{r} - (\Vec{F} \cdot \boldsymbol\omega)\boldsymbol\omega
\end{align}

Thus, for any point $p$ the net force (and thus acceleration) is maximized when $\Vec{F}^*$ is tangent to the circle defined by $\Vec{r}$. Selecting the point $p$ which produces the maximal linear acceleration when $\Vec{F}^*$ is applied there is simply the point $p$ on the child link that maximizes $||\Vec{r}||$, or the point on the object farthest from the axis of rotation $\boldsymbol\omega$. Thus, the optimal policy to articulate a revolute joint is to pick the point on the surface farthest from the axis of rotation $\boldsymbol\omega$ and apply a force parallel to $\Vec{r} \times \boldsymbol\omega$ at every time step.


\subsection{Articulation Parameters to 3D Articulation Flow}

These parameterizations\footnote{An astute reader may recognize that these parameterizations are special cases of twists from Screw Theory. Without loss of accuracy, we choose to omit a rigorous screw-theoretic treatment of articulated objects in favor of an explanation that requires only basic knowledge of physics. } are an elegant representation of single articulations in isolation. However, when an object contains more than one articulation, or contains points that do not move at all (e.g. the base of a cabinet), in order to create a minimal parametric representation of the object we must describe a kinematic tree (a tree of rigid links, connected by joints described by a set of parameters) and associate each point on the object with a link. This is a hierarchical representation, which is difficult to construct from raw observation without prior knowledge of the hierarchical structure or link membership. A hierarchy-free representation of the kinematic properties of the object could assign each point on the object its own set of parameters; however, this would require a full 6 parameters $(v, \omega)$ for each point on the object, and the position $v$ can occur anywhere in $\mathbb{R}^3$ depending on the object's coordinate frame. A more compact, bounded, hierarchy-free representation is 
the \textbf{3D articulation flow (3DAF)} that each point on the object would experience were its part articulated in the positive direction with respect to its articulation parameters. In other words, for each point on each link on the object, define a vector in the direction of motion of that point caused by an infinitesimal displacement $\delta \theta$ of the joint, and normalize it by the largest such displacement on the link. Thus, the 3D articulation flow $f_p$ for point $p \in \mathcal{P}_i$ in link $i$ is: 

\begin{equation} 
\label{eq:flow}
    f_{p} = \begin{cases}
        \Vec{v}, &  \text{if } i \text{ is a prismatic joint} \\
        \frac{\omega \times \Vec{r}}{||\omega \times \Vec{r}_{\mathrm{max}}||} & \text{if } i \text{ is a revolute joint} \\
    \end{cases} 
\end{equation}

where $\Vec{v}, \omega,$ and $\Vec{r}$ are defined above; note that $\Vec{v}$ is already a unit vector. We denote the full set of flow vectors for an object as $F = \{f_p\}_{p \in \mathcal{P}}$ where $\mathcal{P} = \cup_i \mathcal{P}_i $.

While this representation is mathematically equivalent to both the hierarchical and point-wise parameter-based representations, 3D articulation flow has 
several key advantages over parameter-based representations:

\begin{enumerate}
    \item It is hierarchy-free, meaning that it can be easily approximated without an explicit model (i.e. kinematic structure); this property will allow our learned method to generalize to novel object categories.
    \item Each element in the representation is a scaled orientation vector constrained to lie inside the unit sphere in $\mathbb{R}^3$. This means that the representation is invariant under translation and scaling in the coordinate frame of the underlying object. 
\end{enumerate}

Since this representation is defined for any arbitrary point in or on an object, it could be applied to any discrete or continuous geometric representation of said object. However, for the purposes of this work, we apply this representation to 3D point clouds produced from depth images.  Thus for a pointcloud $P = \{p_k \in \mathbb{R}^3\}_{k \in [n]}$, we associate each point $p_k$ in $P$ with a flow vector $f_k \in \mathbb{R}^3, \mathrm{s.t.} ||f_i|| \leq 1$. 


This formulation of 3D articulation flow is similar in spirit to the intermediate representation proposed by \citet{Zeng2020-tk} in their articulation estimation system, FormNet. However, our representation differs in two key ways. First, our representation describes the instantaneous motion of a link, whereas the FormNet formulation predicts the current absolute displacement of a part from a reference position (i.e. a fully-closed door).
Second, we demonstrate that our formulation can be used directly by a manipulation policy, whereas the downstream task of FormNet's representation was predicting the articulation parameters of an object. 


\subsection{Predicting 3D Articulation Flow from Vision}

We now turn to the question of estimating 3D Articulation Flow from a robot's sensor observations. 
We consider a single articulated object in isolation; let $s_0\in \mathcal{S}$ be the starting configuration of the scene with a single articulated object 
where $\mathcal{S}$ is the configuration space. 
We assume that the robot has a depth camera and records point cloud observations $O_t \in \mathbb{R}^{3\times N}$, where $N$ is the total number of observable points from the sensor. 
The task is for the robot to articulate a specified part through its entire range of motion.

For each configuration $s_t$ of the object, there exists a unique ground-truth flow $F_t$, where the ground-truth flow of each point is given by Equation~\ref{eq:flow}. Thus, we would like to find a function $f_{\theta}(O_t)$ that predicts the 3D articulation flow directly from point cloud observations.
We define the objective of minimizing the L2 error of the predicted flow:
\begin{equation}
    \mathcal{L}_{\text{MSE}} = \sum_i{||F_{t,i} - f_\theta(O_t)_i||_2}
\end{equation}
where $i$ indexes over the objects in the training dataset.  While $f_\theta$ can be any estimator, we choose to use a  neural network, which can be trained via a standard supervised learning with this loss function.




\begin{figure*}[ht]
    \centering
    \includegraphics[width=\textwidth]{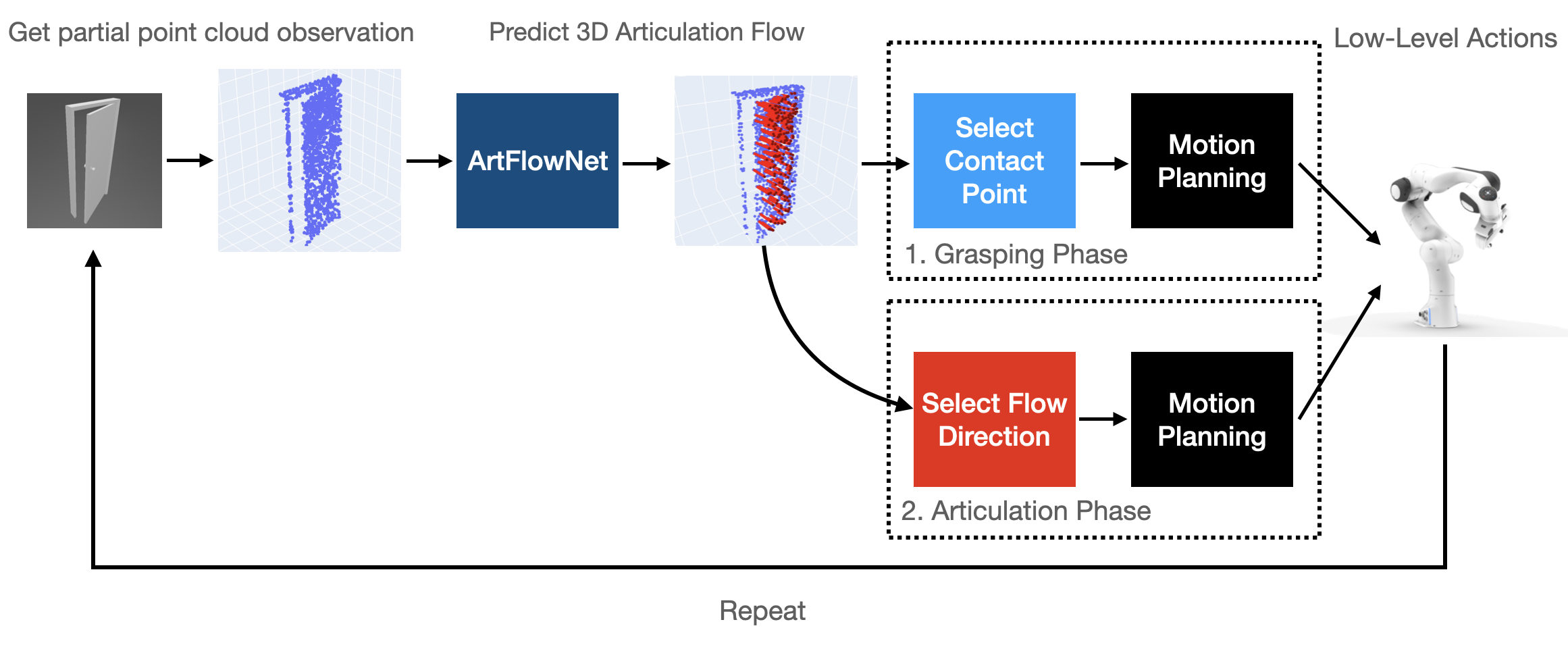}
    \caption{FlowBot3D System Overview. Our system in deployment has two phases: the Grasp-Selection phase and the Articulation-Execution Phase. The dark red dots represent the predicted location of each point, and the light red lines represent the flow vectors connecting from the current time step's points to the predicted points. Note that the flow vectors are downsampled for visual clarity. In Grasp-Selection Phase, the agent observes the environment in the format of point cloud data. The point cloud data will then be post-processed and fed into the ArtFlowNet, which predicts per-point 3D flow vectors. The system then chooses the point that has the maximum flow vector magnitude and deploys motion planning to make contact with the chosen point using suction. In Articulation-Execution phase, after making suction contact with the chosen argmax point, the system iteratively observes the pointcloud data and predicts the 3D flow vectors. In this phase, the motion planning module would guide the robot to follow the maximum observable flow vector's direction and articulate the object of interest repeatedly.}
    \label{fig:overview}
    \vspace*{-15pt}
\end{figure*}


\subsection{A General Policy using 3D Articulation Flow}

\begin{algorithm}

\caption{The FlowBot3D articulation manipulation policy}\label{alg:gap}
\begin{algorithmic}

\REQUIRE $\theta \gets$ parameters of a trained flow prediction network
\STATE $(O_0) \gets$ Initial observation
\STATE $\hat{F_0} \gets f_{\theta}(O_0, [M_0])$, Predict the initial flow
\STATE $g_0 = \texttt{SelectContact}(O_0, \hat{F_0})$, Select a contact pose.
\STATE $in\_contact \gets$ False
\WHILE{not $in\_contact$}
    \STATE Drive an end effector towards $g_0$
    \IF{$\texttt{DetectContact}()$}
        \STATE $in\_contact \gets $ True
        \STATE $\texttt{Grasp}(g_0)$
    \ENDIF
\ENDWHILE

\STATE $done \gets $ False

\WHILE{not $done$}
    \STATE $(O_t) \gets$ Observation
    \STATE $\hat{F_t} \gets f_{\theta}(O_t, [M_t])$, Predict the current flow
    \STATE $v_t \gets \texttt{SelectDirection}()$
    \STATE Apply a force to the end-effector in the direction of $v$ for small duration $t$
    \STATE $done \gets \texttt{EpisodeComplete}()$
\ENDWHILE

\end{algorithmic}
\end{algorithm}


Our method first takes an observation $O_0$ and estimates the 3D articulation flow $\hat{F_0} = f_{\theta}(O_0)$ for all points in the observation. 
Given the estimate of the 3D articulation flow $\hat{F_0}$, we now describe a general, closed-loop policy which takes flow as input and actuates an articulated object. The policy is executed in two phases:

\textbf{1) Grasp Selection}: 
Based on the estimated 3D articulation flow $\hat{F_0}$,
the policy must decide the best place to grasp the object. In this work, we assume access to a suction-type gripper that (in the ideal case) can grasp any point on the object surface.  We know that the ideal attachment point is the location on a part where the flow has the highest magnitude in order to achieve the most efficient actuation of the articulated part by maximizing its acceleration, as we showed by maximizing Equations \ref{eq:prism} and \ref{eq:rot}. We use motion planning to move the end effector to this point,
with the end-effector aligned to directly oppose the flow direction.
We then grasp the object at this position (using a suction gripper), shown in the left hand side of Fig. \ref{fig:overview}.  We assume a rigid contact between the gripper and this contact point 
going forward.

\textbf{2) Articulation Execution}: At each time step $t$, we record a new observation $O_t$ and estimate the current flow $\hat{F}_t$. We then select the predicted flow direction $\Vec{v}_t$ with the greatest magnitude from the visible points from the observation, as shown in the right hand side of Fig. \ref{fig:overview}. To handle objects with multiple articulated parts, we only consider flow vectors close to our point of contact (the contact point itself is likely occluded by the gripper and is thus not visible). 
While continuing to grasp the object, we then move the gripper in the direction $\Vec{v}_t$.
This process repeats in a closed loop fashion until the object has been fully-articulated, a max number of steps has been exceeded, or the episode is otherwise terminated.
See Algorithm \ref{alg:gap} for a full description of the generalized flow articulation algorithm.

\subsection{FlowBot3D: A Robot Articulation System} 

With all the pieces of our generalized articulation policy in place, we now describe a real-world robot system -- FlowBot3D -- which leverages this 
generalized articulation policy. We define a tabletop workspace that includes a Sawyer BLACK 7-DoF robotic arm mounted to the tabletop with a pneumatic suction gripper as its end-effector, and an Azure Kinect RGB-D camera mounted at a fixed position and pointing at the workspace. See Figure \ref{fig:workspace} for an image of the workspace. We obtain point cloud observations of the scene from the Azure Kinect in the robot's base frame,  filtering out non-object points, we use the method proposed in \cite{Zhang2020-xs} to denoise the data (see supplementary materials for details).
For robot control, we use a sampling-based planner, MoveIt! \cite{coleman2014reducing}, which can move our robot to any non-colliding pose in the scene; we thus use motion planning to move the gripper to a pre-grasp pose.  For the grasp and articulation, we directly control the end-effector velocity.




To select the point of contact for the suction gripper, we need to make some modifications from the idealized system described earlier.
Unfortunately, a real suction gripper cannot make a proper seal on locations with high curvature (i.e. edges of the object and uneven surface features such as handles). Since the flow vector with the maximum magnitude is often at one of these extreme points, we must choose an alternative grasp point. While contact selection for suction-based grasping is a well-studied problem \cite{avigal2021avplug, mahler2018dex, mahler2019learning}, we find that a simple heuristic performs acceptably; we choose the point with the highest flow magnitude subject to the following constraints:

\begin{enumerate}
    \item The point itself is not within a certain distance of an edge, where edges are computed using a standard edge-detection algorithm (see supplement for details).
    \item The estimated Gaussian curvature of that point does not exceed a certain threshold (see supplement for details).
    \item The point is not within a distance of $d$ of any points violating conditions 1 and 2.  In practice, we set $d = \texttt{2cm}$ (the radius of the suction tip).
\end{enumerate}
Using this grasp selection method, we are able to execute our general articulation manipulation policy on a real robot. See the supplementary materials for other implementation details.

\subsection{Training Details}

We design a flow prediction network -- which we refer to as ArtFlowNet -- using the dense prediction configuration of PointNet++ \cite{Qi2017-dc} as a backbone, 
and train it using standard supervised learning with the Adam optimizer. 
We emphasize that we train a \textit{single} model to predict 3DAF across all categories, using  a dataset of synthetically-generated (observation, ground-truth flow)
pairs 
based on the ground-truth kinematic and geometric structure provided by the PartNet-Mobility dataset~\cite{Xiang2020-oz}. During each step of training, we select an object in the dataset, randomize the state $S$ of the object, and compute a new supervised pair $(O_S, F_S)$, which we use to compute the loss and update the model parameters. During training, each object is seen in 100 different randomized configurations. Details of our dataset construction and model architecture can be found in the supplementary materials.  

\section{Results}
\label{sec:results}
\begin{table*}[ht]
\renewcommand{\arraystretch}{1.2}
\resizebox{\textwidth}{!}{
\setlength\tabcolsep{.2em}
\begin{tabular}{|r|c|ccccccccccc||c|cccccccccc|}
 \hline
 \multicolumn{13}{|c||}{\textbf{Novel Instances in Train Categories}} 
 &
 \multicolumn{11}{c|}{\textbf{Test Categories}}
 \\
\hline
\rule{0pt}{2.5em}
 & \textbf{\underline{AVG.}} &\clipart{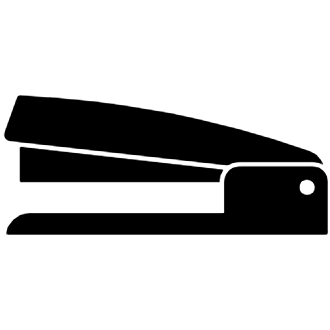} &\clipart{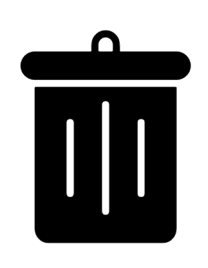}&\clipart{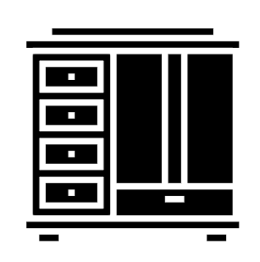}&\clipart{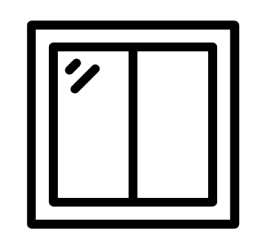}&\clipart{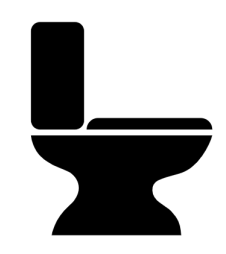}&\clipart{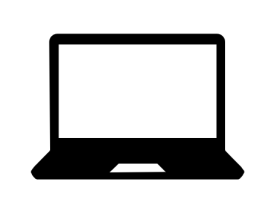}&\clipart{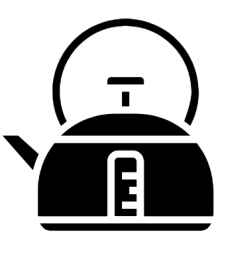}&\clipart{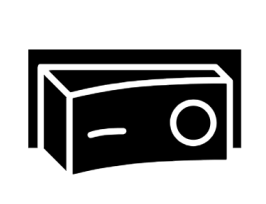}&\clipart{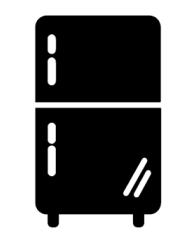}&\clipart{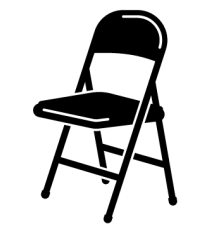}&\clipart{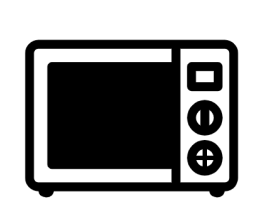}&\textbf{\underline{AVG.}}&\clipart{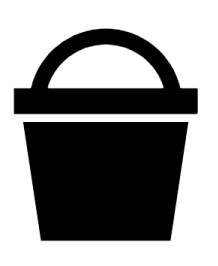}&\clipart{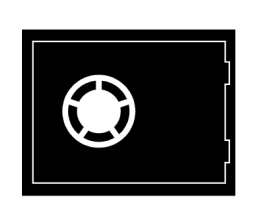}&\clipart{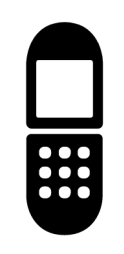}&\clipart{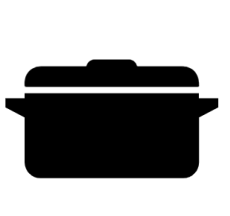}&\clipart{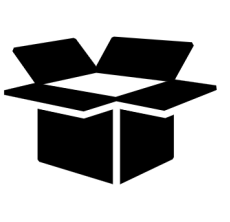}&\clipart{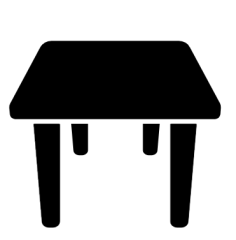}&\clipart{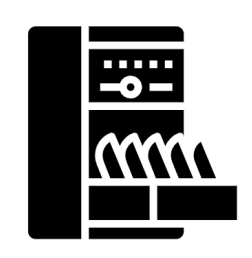}&\clipart{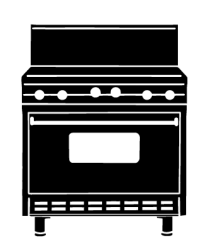}&\clipart{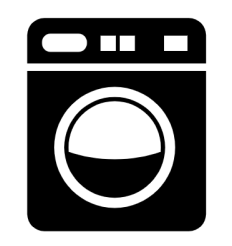}&\clipart{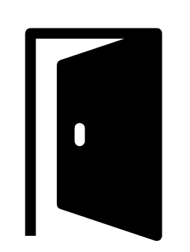}\\
 \hline
 \textbf{Baselines} & \multicolumn{23}{c|}{} \\
 \hline 
 UMP-DI  & 0.29&	\textbf{0.32}&	0.33&	0.16&	0.18&	0.37&	0.14&	0.19&	\textbf{0.28}&	0.72&	\textbf{0.00}&	0.55&0.36&	0.32&	0.62&	0.15&	0.00	&0.41&	0.61&	0.17&	0.34&	0.38&	0.58\\
 Normal Direction&0.40&0.52&0.67&0.16&0.19&0.51&0.60&0.13&0.11&0.55&0.61&0.32&0.39&0.69&0.57&0.04&0.00&0.67&0.19&0.66&1.00&0.43&0.26\\
Screw Parameters&0.40&0.42&0.40&0.42&0.18&0.57&0.45&0.27&0.59&0.51&0.58&0.06&0.18&0.19&0.26&0.08&0.08&\textbf{0.23}&\textbf{0.06}&0.14&0.21&0.34&0.24\\
 BC & 0.74&0.59&0.91&0.63&0.75&0.57&1.00&1.00&0.98&0.62&0.96&0.10&0.87&0.81&0.63&1.00&0.93&0.99&0.74&0.95&0.96&0.83&0.88\\
 DAgger E2E&0.64&0.39&0.85&0.61&0.73&0.50&1.00&0.96&0.90&0.54&0.48&0.10&0.83&0.73&0.62&1.00&0.80&0.95&0.73&0.83&0.98&0.81&0.85\\
 DAgger Oracle&0.51&0.54&0.55&0.20&0.41&0.96&0.64&0.14&0.64&0.47&0.85&0.16&0.56&0.93&0.58&0.61&0.64&0.91&0.27&0.23&0.34&0.27&0.79\\
 \hline
 \textbf{Baselines w/ Flow} & \multicolumn{23}{c|}{} \\
 \hline
 BC + F & 0.83&0.59&1.00&0.61&0.91&1.00&0.97&1.00&1.00&0.69&1.00&0.39&0.91&1.00&0.96&1.00&0.89&0.77&0.71&0.95&0.96&1.00&0.89\\
 DAgger E2E + F&0.76&0.59&0.86&0.60&0.76&0.95&1.00&0.86&0.77&0.65&1.00&0.36&0.91&1.00&0.88&1.00&0.76&0.95&0.68&1.00&0.96&1.00&0.88\\
 DAgger Oracle + F&0.50&0.59&0.53&0.25&0.51&0.58&0.86&0.17&0.65&0.56&0.48&0.38&0.60&0.77&0.71&0.62&0.73&0.91&0.28&0.43&0.47&0.31&0.73\\
 \hline

 \textbf{Ours} & \multicolumn{23}{c|}{} \\
 \hline 
 \textbf{FlowBot3D} & \textbf{0.12}&	{0.32}&	{0.23}&	{0.11}&	{0.09}&	\textbf{0.02}&	\textbf{0.00}&	\textbf{0.09}&	0.32&	\textbf{0.04}&	0.13&	\textbf{0.00}&\textbf{0.15}&	\textbf{0.00}&	{0.21}&\textbf{0.00}&	\textbf{0.00}&	{0.33}&	{0.22}&{0.09}&	{0.07}&\textbf{0.19}&	{0.35}\\
 FlowBot3D w/o Mask&0.17&	0.32&	0.37&	{0.10}&	0.11&	0.15&	0.00&	0.11&	0.33&	0.05&	{0.07}&	0.29&0.19&	0.16&	0.24&	0.05&	0.00&	{0.24}&	0.24&	0.17&	0.27&	0.19&	0.37\\
  FlowBot3D w/o Mask (+VPA) &0.16&	{0.33}&	\textbf{0.09}&	\textbf{0.07}&	\textbf{0.07}&	0.16&	0.00&	0.14&	0.49&	0.27&	0.11&	0.00&0.16&	0.11&	\textbf{0.17}&	0.23&	0.00&	0.53&{0.10}&	\textbf{0.05}&	\textbf{0.00}&	0.23&	\textbf{0.20}\\
  \hline
  Oracle w/ GT 3DAF&0.05&0.10&0.10&0.03&0.11&0.06&0.00&0.12&0.00&0.00&0.02&0.00&0.16&0.00&0.12&0.95&0.00&0.12&0.14&0.02&0.00&0.13&0.12\\
  \hline
\end{tabular}
}
\smallskip

\caption{Normalized Distance Metric Results ($\downarrow$): Normalized distances to the target articulation joint angle after a full rollout across different methods. The lower the better.}
\label{tab:baselines-dist}
\end{table*}

\begin{table*}[ht]
\renewcommand{\arraystretch}{1.2}
\resizebox{\textwidth}{!}{
\setlength\tabcolsep{.2em}
\begin{tabular}{|r|c|ccccccccccc||c|cccccccccc|}
 \hline
 \multicolumn{13}{|c||}{\textbf{Novel Instances in Train Categories}} 
 &
 \multicolumn{11}{c|}{\textbf{Test Categories}}
 \\
\hline
\rule{0pt}{2.5em}
 & \textbf{\underline{AVG.}} &\clipart{stapler} &\clipart{trash}&\clipart{storage}&\clipart{window}&\clipart{toilet}&\clipart{laptop}&\clipart{kettle}&\clipart{switch}&\clipart{fridge}&\clipart{chair}&\clipart{microwave}&\textbf{\underline{AVG.}}&\clipart{bucket}&\clipart{safe}&\clipart{phone}&\clipart{pot}&\clipart{box}&\clipart{table}&\clipart{dish}&\clipart{oven}&\clipart{washer}&\clipart{door}\\
 \hline
 \textbf{Baselines} & \multicolumn{23}{c|}{} \\
 \hline 
   UMP-DI   & 0.52&	{0.60}&	0.33&	0.65&	0.73&	0.29&	0.67&	0.80&	\textbf{0.50}&	0.11&	\textbf{1.00}&	0.00&0.45&	0.83&	0.03&	0.50&	1.00&	0.31&	0.29&	0.78&	0.33&	0.31&	0.20\\
 Normal Direction&0.31&0.40&0.00&0.51&0.71&0.00&0.00&0.80&0.50&0.00&0.00&0.50&0.31&0.00&0.00&0.50&1.00&0.00&0.55&0.00&0.00&0.10&0.64\\
 Screw Parameters&0.50&0.50&0.53&0.51&0.80&0.21&0.55&0.60&0.17&0.37&0.43&0.80&0.67&0.17&0.63&0.67&0.92&\textbf{0.69}&\textbf{0.92}&0.83&0.75&0.50&0.72
\\
   BC & 0.14&0.40&0.00&0.20&0.18&0.14&0.00&0.00&0.00&0.11&0.00&0.50&0.04&0.17&0.00&0.00&0.04&0.00&0.15&0.00&0.00&0.00&0.00\\
 DAgger E2E&0.14&0.60&0.00&0.26&0.09&0.28&0.00&0.00&0.00&0.00&0.25&0.00&0.04&0.00&0.00&0.00&0.20&0.00&0.17&0.02&0.00&0.00&0.00\\
 DAgger Oracle&0.29&0.40&0.33&0.80&0.27&0.00&0.00&0.80&0.00&0.11&0.00&0.50&0.20&0.00&0.00&0.00&0.36&0.00&0.29&0.49&0.33&0.31&0.20\\
 \hline
 \textbf{Baselines w/ Flow} & \multicolumn{23}{c|}{} \\
 \hline
 BC + F & 0.11&0.40&0.00&0.26&0.10&0.00&0.00&0.00&0.00&0.00&0.00&0.50&0.03&0.00&0.00&0.00&0.08&0.00&0.15&0.05&0.00&0.00&0.00\\
 DAgger E2E + F&0.14&0.40&0.00&0.00&0.09&0.00&0.00&0.00&0.50&0.00&0.00&0.50&0.04&0.00&0.00&0.00&0.20&0.00&0.15&0.05&0.00&0.00&0.00\\
 DAgger Oracle + F&0.33&0.40&0.33&0.58&0.45&0.00&0.00&0.80&0.00&0.11&0.50&0.50&0.16&0.00&0.00&0.00&0.24&0.00&0.34&0.41&0.33&0.12&0.16\\
 \hline

 \textbf{Ours} & \multicolumn{23}{c|}{} \\
 \hline 
 \textbf{FlowBot3D} & \textbf{0.77}&	0.57&	{0.56}&	{0.88}&	{0.82}&	\textbf{0.86}&	\textbf{1.00}&	\textbf{0.80}&	{0.50}&	{0.78}&	{0.75}&	\textbf{1.00}&{0.69}&	\textbf{1.00}&{0.56}&	\textbf{1.00}&	\textbf{1.00}&	{0.38}&	{0.43}&{0.84}&{0.83}&{0.43}&{0.44}\\
 FlowBot3D w/o Mask&0.72&	\textbf{0.67}&	0.55&	0.85&	0.82&	0.57&	1.00&	0.80&	0.50&	\textbf{0.89}&	{0.75}&	0.50&0.62&	0.83& {0.59}&	0.50&	1.00&	{0.62}&	0.28&	0.76&	0.58&	{0.56}&	{0.50}\\

 FlowBot3D w/o Mask + VPA&0.73&	0.60&	\textbf{0.67}&	\textbf{0.88}&	\textbf{0.91}&	0.43&	1.00&	0.80&	0.50&	0.56&	0.72&	1.00&\textbf{0.70}&	0.50&	\textbf{0.63}&	0.50&	1.00&	0.23&	{0.90}&	\textbf{0.88}&	\textbf{1.00}&	\textbf{0.63}&	\textbf{0.72}\\
  \hline
  Oracle w/ GT 3DAF&0.92&0.80&1.00&0.97&0.82&0.71&1.00&0.80&1.00&1.00&1.00&1.00&0.82&1.00&0.85&0.00&1.00&0.85&0.86&0.98&1.00&0.81&0.88\\
  \hline
\end{tabular}
}
\smallskip

\caption{Success Rate Metric Results ($\uparrow$): Fraction of success trials (normalized distance less than 0.1) of different objects' categories after a full rollout across different methods. The higher the better. }
\vspace*{-10pt}
\label{tab:baselines-sr}
\end{table*}

\begin{figure*}[!htb]
\vspace*{10pt}
\centering
\begin{minipage}[b]{.47\textwidth}
    \centering
    \includegraphics[width=\linewidth]{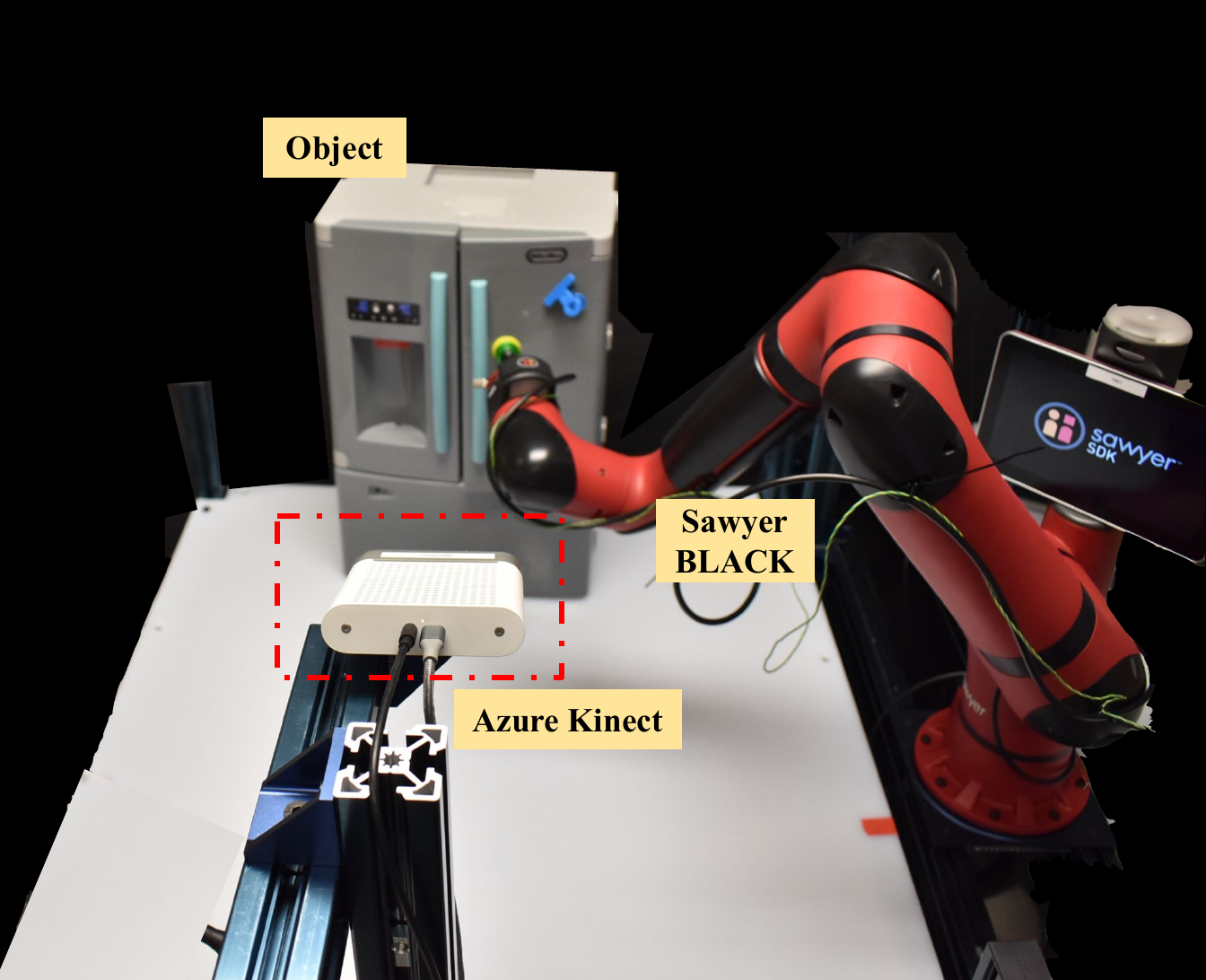}
    \caption{Workspace setup for physical experiments. The sensory signal comes from an Azure Kinect depth camera, and the agent is a Sawyer BLACK robot.}
    \label{fig:workspace}
\end{minipage}\qquad
\begin{minipage}[b]{.47\textwidth}
    \centering
    \includegraphics[width=\linewidth]{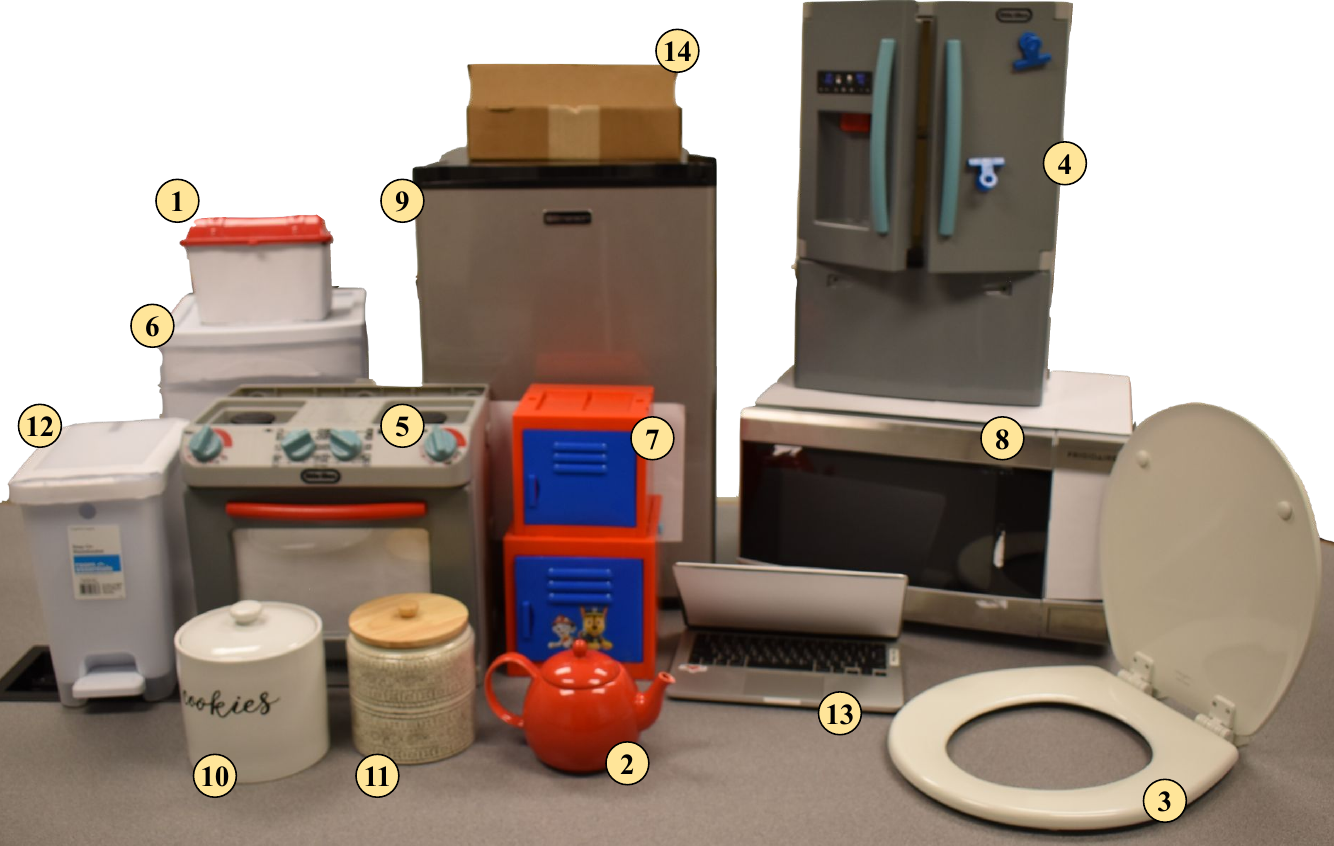}
    \caption{Fourteen test objects for our real-world experiments. Please refer to Supplementary Material for the exact category of each object.}
    \label{fig:all_rw}

\end{minipage}
        \vspace*{-15pt}

\end{figure*}

We conduct a wide range of simulated and real-world experiments to evaluate the FlowBot3D system.

\subsection{Simulation Results}
To evaluate our method in simulation, we implement a suction gripper in the ManiSkill environment \cite{Mu2021-ui}, which serves as a simulation interface for interacting with the PartNet-Mobility dataset \cite{Xiang2020-oz}. The PartNet-Mobility dataset contains 46 categories of articulated objects; following UMPNet~\cite{Xu2021-iw}, we consider a subset of PartNet-Mobility containing 21 classes, split into 11 training categories (499 training objects, 128 testing objects) and 10 entirely unseen object categories (238 unseen objects). Several objects in the original dataset contain invalid meshes, which we exclude from evaluation. We  modify ManiSkill simulation environment to accommodate these object categories. We train our models (ArtFlowNet and baselines) exclusively on the training instances of the training object categories, and evaluate by rolling out the corresponding policies for every object in the ManiSkill environment. Each object starts in the ``closed'' state (one end of its range of motion), and the goal is to actuate the joint to its ``open'' state (the other end of its range of motion). For experiments in simulation, we include in the observation $O_t$ a binary part mask indicating which points belong to the child joint of interest. Results are shown in Tables \ref{tab:baselines-dist} and \ref{tab:baselines-sr}\footnote{Categories from left to right: stapler, trash can, storage furniture, window, toilet, laptop, kettle, switch, fridge, folding chair, microwave, bucket, safe, phone, pot, box, table, dishwasher, oven, washing machine, and door. Clipart pictures are borrowed from UMPNet paper with the authors' permission.
}.


\textbf{Metrics. } During our experiments, we calculate two metrics:
\begin{itemize}
    \item \emph{Normalized distance}: Following \citet{Xu2021-iw}, we compute the normalized distance travelled by a specific child link through its range of motion. The metric is computed based on the final configuration after a policy rollout ($\Vec{j}_\mathrm{end}$) and the initial configuration ($\Vec{j}_\mathrm{init}$):
\[
\mathcal{E}_\mathrm{goal} = \frac{||\Vec{j}_\mathrm{end} - \Vec{j}_\mathrm{goal}||}{||\Vec{j}_\mathrm{goal} - \Vec{j}_\mathrm{init}||}
\]

\item \emph{Success}: We also define a binary success metric, which is computed by thresholding the final resulting normalized distance at $\delta$:
$
\mathrm{Success} = \mathbbm{1}(\mathcal{E}_\mathrm{goal} \leq \delta)
$.
We set $\delta = 0.1$, meaning that we define a success as articulating a part for more than 90\%. 

\end{itemize}

\textbf{Baseline Comparisons}: We compare our proposed method with several baseline methods:
\begin{itemize}

    \item \textbf{UMP-DI}: We implement a variant\footnote{We could not yet compare directly to UMPNet, as  their model and simulation environment had not yet been released at the time between submission and publication.} of UMPNet's Direction Inference network  (DistNet) \cite{Xu2021-iw}, where instead of bootstrapping an action scoring function from interaction, we learn the scoring function by regressing the cosine distance between a query vector and the ideal flow vector for a contact point. At test time, we select the contact point based on ground-truth 3DAF, and after contact has been achieved we use CEM to optimize the scoring function to predict the action direction at every timestep.
    \item \textbf{Normal Direction}: We use off-the-shelf normal estimation to estimate the surface normals of the point cloud using Open3D \cite{zhou2018open3d}. To break symmetry, we align the normal direction vectors to the camera. At execution time, we first choose the ground-truth maximum-flow point and then follow the direction of the estimated normal vector of the surface.
    
    \item \textbf{Screw Parameters}: \hl{We predict the screw parameters for the selected joint of the articulated object. We then generate 3DAF from these predicted parameters and use the FlowBot3D policy on top of the generated flow.} 
    
    \item \textbf{Behavioral Cloning (BC)}: The agent takes as input a point cloud and outputs the action of the robot. The agent uses the PointNet-Transformer architecture proposed in \cite{Mu2021-ui}. The agent is trained end-to-end via L2 regression on trajectories provided by an oracle version of GT 3DAF.
    
    \item \textbf{BC + F}: Same as BC, but with ground-truth flow at input.
    
    \item \textbf{DAgger E2E}: We also conduct behavioral cloning experiments with DAgger \cite{ross2011reduction} on the same expert dataset as in the BC baseline. We train it end-to-end (E2E), similar to the BC model above.
    
    \item \textbf{DAgger E2E + F}: Same as DAgger E2E, but with ground-truth flow as an input.

    \item \textbf{DAgger Oracle}:  A two-step policy, where we first use ground-truth flow to select a contact point using the Generalized Articulation Policy heuristic, and train DAgger on expert trajectories generated after the point of contact.
    \item \textbf{DAgger Oracle + F}: Same as DAgger Oracle, but with ground-truth flow at input.
    \item \textbf{Oracle w/ GT 3DAF}: An oracle version of FlowBot3D that uses ground truth 3DAF vectors instead of the predicted ones for both phases. This serves as an upper bound of FlowBot 3D's performance.\footnote{The description files of the phone meshes contain wrong rotation axis, thus the poor performance of the oracle policy on that category.}

\end{itemize}
Each method above consists of a single model trained across all  PartNet-Mobility training categories. For a more straightforward comparison, we dedicate Table \ref{tab:baselines-dist} and \ref{tab:baselines-sr} to evaluations in the SAPIEN simulator and we defer the comparison between UMPNet and FlowBot3D to the supplementary material.

\textbf{Analysis}: We can draw two conclusions from our simulated evaluation. First, our formulation of FlowBot3D has a very high success rate across all categories, including test categories, which are completely novel types of objects (but may contain similar parts and articulation structures). This is evidence that the ArtFlowNet network is learning salient geometric features to predict the location and character of articulated points. Based on visual interpretation of actual predicted flows, ArtFlowNet is particularly adept at recognizing doors, lids, drawers, and other large articulated features. One might have thought that 3DAF is essentially estimating normal directions, but this is not the case, as seen in the results of the \hl{Normal Direction baseline. Normal Direction estimation suffers from occlusion issues and the normal is not always the correct direction to actuate the object (for example, for the spherical-shaped lid of a teapot)}. Additionally, our method's accuracy increases when the object is at least partially open, because there is less ambiguity about object structure than when an object is fully ``closed''. The UMP-DI baseline exhibits similar properties, but the implicit optimization yields noisier direction predictions. Second, none of the Behavior Cloning and DAgger policies, nor their flow-based variants, perform well. The best BC baseline, DAgger Oracle + F, is only able to fully articulate objects 33\% of the time.


\textbf{UMPNet Pybullet Environment}: \hl{The simulation environment used in the original UMPNet evaluations} \cite{Xu2021-iw} \hl{is a PyBullet-based environment with different physical and collision parameters. However, the source code to run the UMPNet environment was not available for us to run until after this paper was submitted for review; we have since obtained a copy of this environment, and evaluate our method on their environment in the supplementary materials.}

\subsection{Real-World Experiments}

\begin{figure*}[ht]
    \centering
    \includegraphics[width=\linewidth]{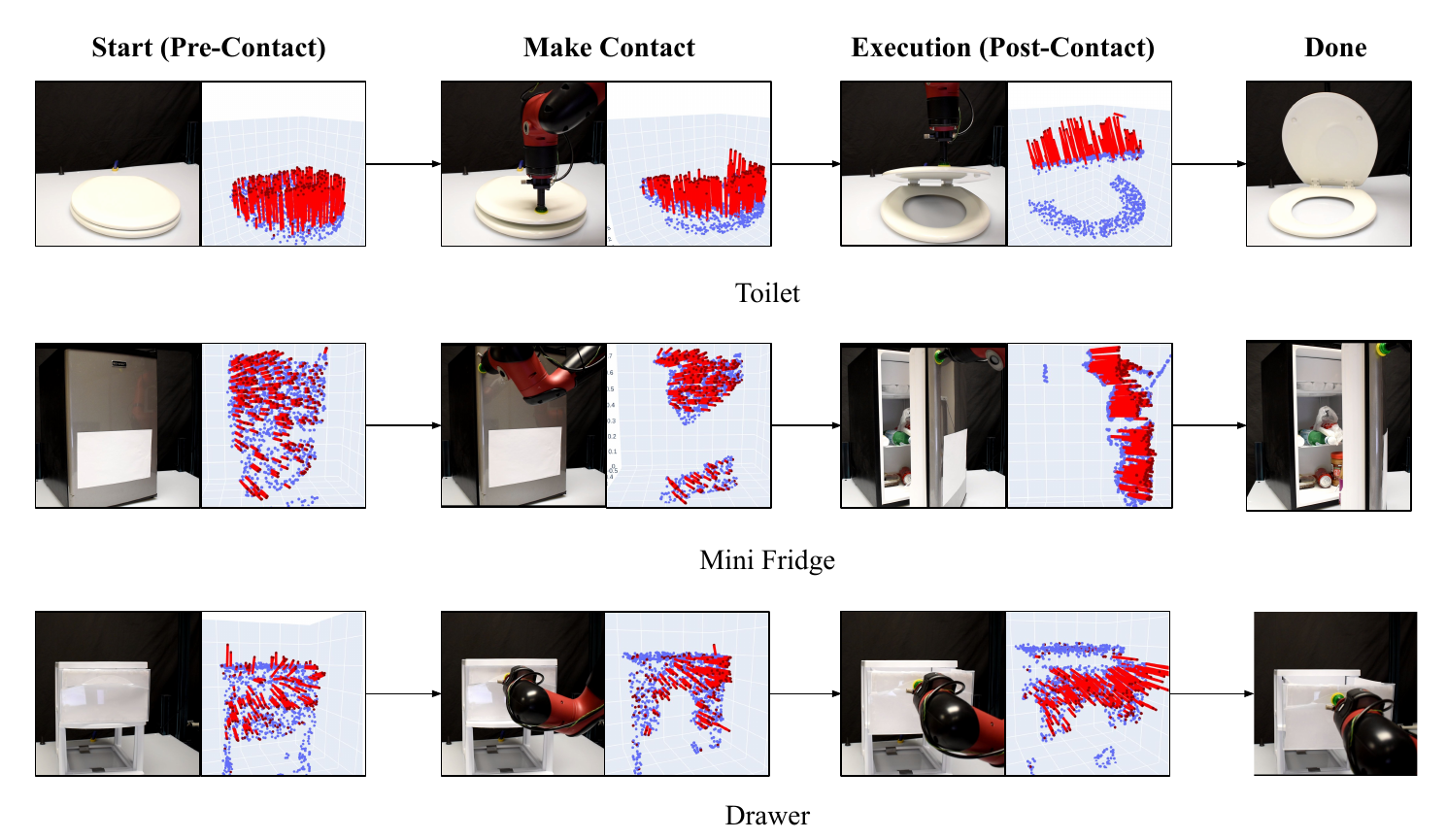}
    \vspace*{-15pt}
    \caption{Real world examples of FlowBot3D executing an articulation policy based on predicting 3D Articulated Flow. Notice that even with occlusions, such as in the intermediate mini-fridge observation, the network is able to predict reasonable 3D articulation flow vectors for downstream policy.}
    \label{fig:real_world_results}
    \vspace*{-10pt}
\end{figure*}

To evaluate the performance of FlowBot3D when executed in a real robotic environment, we design a set of of real-world experiments in which we attempt to articulate a variety of different household objects using the Sawyer robot in our workspace, as shown in Fig. \ref{fig:workspace}. Our experiment protocol is thus: for each object in the dataset, we conducted 5 trials of each method. For each trial, the object is placed in the scene at a random position such that the articulations are visible and the robot can reach every position in the range of motion of each articulation. The policy is then executed for at most 10 steps, terminating earlier if success has been achieved or if the policy predicts an action that cannot be executed safely (this case is infrequent). We conducted one round of evaluation (70 trials in total) for each of the following methods:

\begin{itemize}
    \item \textbf{FlowBot3D}: The version of our generalized articulation policy as presented in Section \ref{sec:method}. In experiments, we use an ArtFlowNet trained without a part mask in the observation space. In addition, since the camera position in reality is different from that in the ManiSkill environments, we apply a viewpoint augmentation (VPA) at training time, where we render synthetic point clouds from various camera angles in simulation.
    \item \textbf{DAgger Oracle}: The DAgger model trained in simulation to produce closed-loop motion directions, but with the contact selection predicted by the FlowBot3D model.
\end{itemize}
As in our simulated experiments, we use a single model trained in simulation across multiple object categories without any further finetuning.


\begin{table}[ht]
    \centering
    \renewcommand{\arraystretch}{1.2}
    \setlength\tabcolsep{.16em}    \begin{tabular}{|c|ccccccccccc|}
        \hline
        \rule{0pt}{2.5em}

         & \clipart{box} & \clipart{kettle} & \clipart{toilet} & \clipart{fridge} & \clipart{oven} & \clipart{storage} & \clipart{safe} & \clipart{microwave} & \clipart{laptop} & \clipart{trash}  & \clipart{pot} \\
         \hline
         \# Objects & 2 & 1 & 1 & 2 & 1 & 1 & 1 & 1 & 1 & 1 & 2 \\
         \hline
    \end{tabular}
    \caption{Real-world objects used during our experiments.}
    \vspace*{-5pt}
    \label{tab:summary}
\end{table}

\textbf{Objects}: 
We assemble a set of real-world objects that are representative of typical articulations a human may encounter in the real world: doors, drawers, hinges, etc. The objects were selected before experimentation began, and the only criteria for inclusion were 1) that it fit in the workspace, 2) it had a surface that a suction gripper could attach to and actuate, and 3) it wasn't too dark or reflective, so as to be seen by the Azure Kinect's depth camera. Each object falls into one of either the training or test classes we selected from the PartNet-Mobility. We also include several jars with lids, which, while not strictly articulated as 1-DoF joints, can be articulated like a prismatic joint. See Figure \ref{fig:all_rw} and Table \ref{tab:summary} for a summary of the dataset, and the supplementary materials for specifics for each object.

\textbf{Metrics}: During our trials, we compute the following metrics for each policy:

\begin{itemize}
    \item \underline{Overall Success}: Was the object articulated more than 90\% of its range of motion (defined per-object)?
    \item \underline{Contact Success}: Was the contact point chosen on an a joint that can move, and was the suction tip able to successfully form a seal at that point?
    \item \underline{Average Distance}: Conditioned on a successful contact, what was the average distance from the end of the object's range of motion after the policy terminated?
    \item \underline{Motion Success}: After successful contact, was the object articulated more than 90 \% of its range of motion?
\end{itemize}

Details about how our trials are conducted and measurements computed can be found in the supplementary materials.

\begin{table}[]
\centering
\setlength\tabcolsep{.4em}
\renewcommand{\arraystretch}{1.2}

\resizebox{\columnwidth}{!}{
\begin{tabular}{|c|cccc|}
 \hline
Method & Overall Succ. & Contact Succ. & Avg. Dist. & Motion Succ. \\
 \hline
 \textbf{FlowBot3D} & \textbf{45/70 (64.3\%)} & 64/70 (91.4\%)* & \textbf{0.22} & \textbf{45/64 (70.3\%)} \\
 DAgger Oracle & 10/70 (14.3\%) &\textbf{68/70 (97.1\%)*} & 0.73 & 10/68 (14.7\%) \\
 \hline
\end{tabular}
}
\smallskip
\caption{Trials for FlowBot3D. *Note that both methods in the Contact Success column use the same FlowBot3D contact prediction and execution policy.}
\vspace*{-5pt}
\label{tab:real-world}
\end{table}


\textbf{Quantitative analysis}: We present summary metrics in Table \ref{tab:real-world}, and a per-object summary in our supplementary materials. Across all metrics, FlowBot3D performs substantially better than the DAgger baseline. In absolute terms, the policy succeeds a high fraction of the time (64\%); the policy selects a suitable contact point on the object 91\% of the time, and succeeded 70\% of the time after contact was established.

In contrast, the baseline policy succeeded in a very small number of cases, only 14\% of the time. While contact rates were comparable to the trials conducted for FlowBot3D (they use the same contact selection method), the motions predicted were almost always unsuccessful.

\textbf{Qualitative analysis of FlowBot3D}: A major goal of our real-world trials was to evaluate how well the Flowbet3D policy transfers from simulation to reality without any retraining. We find that the overall policy performs surprisingly well, and the ArtFlowNet module -- trained exclusively on point clouds rendered in simulation -- generalizes impressively to real-world objects, producing high-fidelity flow predictions on a range of real objects. This is because there isn't much of a domain shift in the point cloud observations, and the geometric features that signal an articulation are fairly consistent.



\textbf{Failure modes:} We have found that the majority of trial failures were due to two reasons in real world: flow prediction error and contact failure. For flow prediction errors, after making contact with the object, executing an incorrect 3D articulation flow vector will drive the gripper away from the object, causing the gripper to lose contact with the object. The bulk of flow prediction errors happen either because the robot occludes too much of the scene (which might be rectified by multiple viewpoints, temporal filtering, or a recurrent policy), or because the robot fails to detect the presence of articulations (this occurs on the real oven, for instance). For contact failures, the contact selection heuristic might not filter out all ungraspable points and thus the robot might choose a contact point that is difficult or impossible to make a complete seal on during suction. Overall, we theorize that many of the failures could be mitigated by improving the compliance and control of the gripper, and including a stronger contact quality prediction module.


\subsection{Simulation Ablations}
We conduct two ablations on the design decisions made for ArtFlowNet:

\begin{itemize}
    \item \textbf{Including a part mask} We study the effect of providing a segmentation mask of the articulated part of interest as input to ArtFlowNet; such a mask could theoretically be obtained by a segmentation method or provided by a human to specify the articulation task, as in the SAPIEN challenge \cite{Xiang2020-oz}. We find while the inclusion of a mask improves predictions in ambiguous cases (i.e. when a door is closed and coplanar with its parent link), removing the mask only decreases performance a small amount (see Tables \ref{tab:baselines-dist} and \ref{tab:baselines-sr}).
    \item \textbf{Applying viewpoint augmentations during training}: We analyze how randomizing the camera viewpoint during the synthetic dataset generation step affects model performance. We find that augmenting the viewpoint has little effect on the performance in a simulated environment (see Tables \ref{tab:baselines-dist} and \ref{tab:baselines-sr}), but improves performance in sim-to-real transfer.
\end{itemize}




\section{Conclusion} 
\label{sec:conclusion}
In this work, we propose a novel visual representation for articulated objects, namely \textbf{3D Articulation Flow}, as well as a policy -- \textbf{FlowBot3D} -- which leverages this representation to successfully manipulate articulated objects. We demonstrate the effectiveness of our method in both simulated and real environments, and observe strong sim-to-real transfer generalization.

While our method shows strong performance on a range of object classes, there is substantial room for improvement. One class of improvements is in system-building and engineering: with a more compliant robotic arm controller, as well as a more sophisticated contact prediction system, we believe we would be able to eliminate a wide class of failure modes. However, the remaining failure modes raise questions we would like to explore in future work. For instance, we would like to explore how our flow representation models might be used in an online adaptation setting, so that incorrect predictions can be corrected. We also would like to explore how our representation might be useful when learning from demonstrations, or in other more complex manipulation settings.

\section*{Acknowledgments}

{\footnotesize
This material is based upon work supported by the National Science Foundation under Grant Nos. IIS-1849154 (NSF S\&AS), IIS-2046491 (NSF CAREER), and DGE-1745016 (NSF GRFP), as well as LG Electronics. We are grateful to Thomas Weng, Brian Okorn, Daniel Seita, Shikhar Bahl, and Russell Mendonca for feedback on the paper.
}


\bibliographystyle{plainnat}
\bibliography{references}

\newpage
\appendix



\section{Robot System Details}

\subsection{Hardware}
In all of our real-world experiments, we deploy our system on a Rethink Sawyer Robot and the sensory data (point cloud) come from an Azure Kinect depth camera. The robot's end effector is an official Saywer Pneumatic Suction Gripper with a suction cup with a diameter of 3~cm. The air supply of the suction gripper is provided by a California Air Tools compressor.

\subsection{Workspace}
We set up our workspace in a 1.08~m by 1.00~m space put together using Vention beams. We set up the Azure Kinect camera such that it points toward the center of workspace and has minimal interference with the robot arm-reach trajectory. Collision geometry are set up using MoveIt's collision box construction tool. We add a number of boxes representing the camera and Vention beams that can potentially be blocking the robot during motion planning.

\subsection{Hand-Eye Calibration}
For Hand-Eye Calibration, we are using the Easy-Hand-Eye ROS package that calculates the transformation from the camera frame to world frame using an ArUco marker fixed on the robot's end effector. The process requires about 30 samples of the robot pose and ArUco marker pose combinations. 

\subsection{Foreground Segmentation}
In simulated experiments, we have access to segmentation masks that segment out tabletop and robot from the collected point cloud. In real-world experiments, however, we need to programatically segment out those points ourselves.

\textbf{Tabletop.} We segment out the tabletop plane by simply subtracting the points with $z$ values less than 0.015~m from the collected point cloud after calibration because the table top is placed 1.5~cm below the robot base. 

\textbf{Robot. }The robot points are masked out in real-time by rendering the robot 3D model using its URDF file. This is done through a ROS package called Real Time URDF Filter. This filter assumes a perfect calibration of the camera. When the calibration is slightly off, some trailing points from the robot might remain in scene. Thus, we also statistically remove the outliers from the resulting point cloud because the remaining robot points are sparser than the object's points. 

\subsection{Contact Point Heuristic}
In simulation, the suction contact is modeled by a kinematic constraint between the gripper point and the contact point. Therefore, in simulation, we have a perfect contact that can almost always successfully grasp the desired part. In real-world experiments, due to the complication of the physics of the suction gripper and the geometry of the target part, we can not always guarantee a successful grasp. Therefore, we add an extra heuristic upon the max-flow selection when selecting which point to grasp. Specifically, we add an interior point selection procedure that calculates the curvature of the points using PCA and we choose the point with curvature value smaller than a threshold. If the max-flow point has a curvature value higher than the threshold, we discard that point and choose the nearest low-curvature point at least 2~cm away from the max-flow point.

\subsection{Grasp Selection Details}
In the Grasp Selection phase of real-world experiments, we predict and estimate the part's 3D articulated flow vectors using FlowNet. We then use the aforementioned contact point heuristic to filter out points that have high curvature values. If the max-flow point is within the remaining points, we keep it and use it as the selected contact point. Otherwise, we choose the nearest low-curvature point at least 2~cm away from the max-flow point. Once we have selected the point, we have also selected the end effector's goal translation. For goal orientation, we align the end effector with the flow vector. The procedure is explained here: assume that the axis connecting the suction gripper tip to the robot hand is called $\Vec{v}_1$ and the chosen flow vector is $-\Vec{v}_2$, we aim to find a rotation that aligns $\Vec{v}_1$ to $\Vec{v}_2$ (because the robot approach direction is opposite to the flow direction). The difference of the rotation expressed in quaternion is calculated as follows:
\begin{align*}
    \phi_{12} &= \cos^{-1}(\Vec{v}_1\cdot\Vec{v}_2)\\
    \omega &= \Vec{v}_1\times\Vec{v}_2 = [\omega_x, \omega_y, \omega_z]\\
    q_x&= \omega_x\cdot\sin(\omega/2)\\
    q_y&= \omega_y\cdot\sin(\omega/2)\\
    q_z&= \omega_z\cdot\sin(\omega/2)\\
    q_w&= \cos(\omega/2)\\
    \Vec{q} &= [q_x, q_y, q_z, q_w] \\
    \Vec{q}_{\mathrm{diff}} &= \frac{\Vec{q}}{||\Vec{q}||}.
\end{align*}
Therefore, when given the robot's starting rotation quaternion $\Vec{q}_\mathrm{start}$, the goal orientation of the robot end-effector $\Vec{q}_\mathrm{goal}$ is given by:
\[
\Vec{q}_\mathrm{goal} = \Vec{q}_\mathrm{diff}\cdot \Vec{q}_\mathrm{start}
\]

\subsection{Robot Control Paradigm}
In the Grasp Selection Phase of real-world experiments, the robot is controlled using position control by inputting the end-effector pose and solving for the trajectory using an RRTConnect-based [17] IK solver. One caveat about the Grasp Selection Phase in real world is that the robot does not make contact with the select point directly. Instead, the robot first aligns with the chosen flow vector and plans to a point 10~cm in the chosen 3D articulated flow direction away from the max-flow point. Then the robot switches the control mode to velocity control and approaches the proposed point in the aligned (negative selected flow) direction slowly until the force sensor of the robot reports reading greater than a threshold, meaning the robot makes contact with the object. Then in the Articulation-Execution Phase, the velocity controller takes as input the translational velocity represented by the current time step's normalized predicted articulation flow vector multiplied by a constant to decrease the speed and the rotational velocity as the aforementioned $\Vec{q}_\mathrm{diff}$ converted to Euler angles multiplied by another constant to decrease the angular speed.

\section{Training Details}
\subsection{Network Architecture}
ArtFlowNet is based on the PointNet++ [27] architecture. The architecture largely remains similar to the original architecture except for the output head. Instead of using a segmentation output head, we use a regression head. The ArtFlowNet architecture is implemented using Pytorch-Geometric [10], a graph-learning framework based on PyTorch. Since we are doing regression, we use standard L2 loss optimized by an Adam optimizer [16].
\subsection{Ground Truth 3DAF Generation}
We implement efficient ground truth 3D Articulation Flow generation. At each timestep, the system reads the current state of the object of interest in simulation as an URDF file and parses it to obtain a kinematic chain. Then the system uses the kinematic chain to analytically calculate each point's location after a small, given amount of displacement. In simulation, since we have access to part-specific masks, the calculated points' location will be masked out such that only the part of interest will be articulated. Then we take difference between the calculated new points and the current time step's points to obtain the ground truth 3D Articulation Flow. 
\subsection{Simulator Modifications}
We heavily modify the ManiSkill [23] environment, which is a high-level wrapper of the SAPIEN [32] simulator. Specifically, we add in a variety of PartNet-Mobility objects to the environment for more diverse training dataset. We obtain a list of training and testing objects from the authors of UMPNet [33]. We have filtered out some phone objects and door objects due to the collision of meshes in the SAPIEN simulator upon loading, but the dataset remains largely identical to the one used in UMPNet. Furthermore, we implement efficient online ground truth 3D articulation flow calculation in the ManiSkill environment for generating training data online. We also implement camera viewpoint sampling by randomizing the azimuth and elevation for the VPA model training. Instead of using a full robot arm, we only use a floating gripper with 8~DoF ($x, y, z$ for translation, $r, p, y$ for rotation, and speed parameter for each of the two fingers on the gripper) controlled by a velocity controller. The two gripper fingers' speed parameters are not learned in Behavioral Cloning as the two fingers remain closed. To simulate suction, we create a strong force between the gripper fingers and the target object since kinematic constraints are not directly supported in the SAPIEN simulator.
\subsection{Hyperparameters}
We use a batch size of 64 and a learning rate of 1e-4. We use the standard set of hyperparameters from the original PointNet++ paper.

\section{Simulation Experiments Illustration}
Here we briefly illustrate FlowBot3D system in simulation. In simulation, the suction is implemented using a strong force between the robot gripper and the target part. 

\begin{figure*}[htb]
\centering
\includegraphics[width=\linewidth]{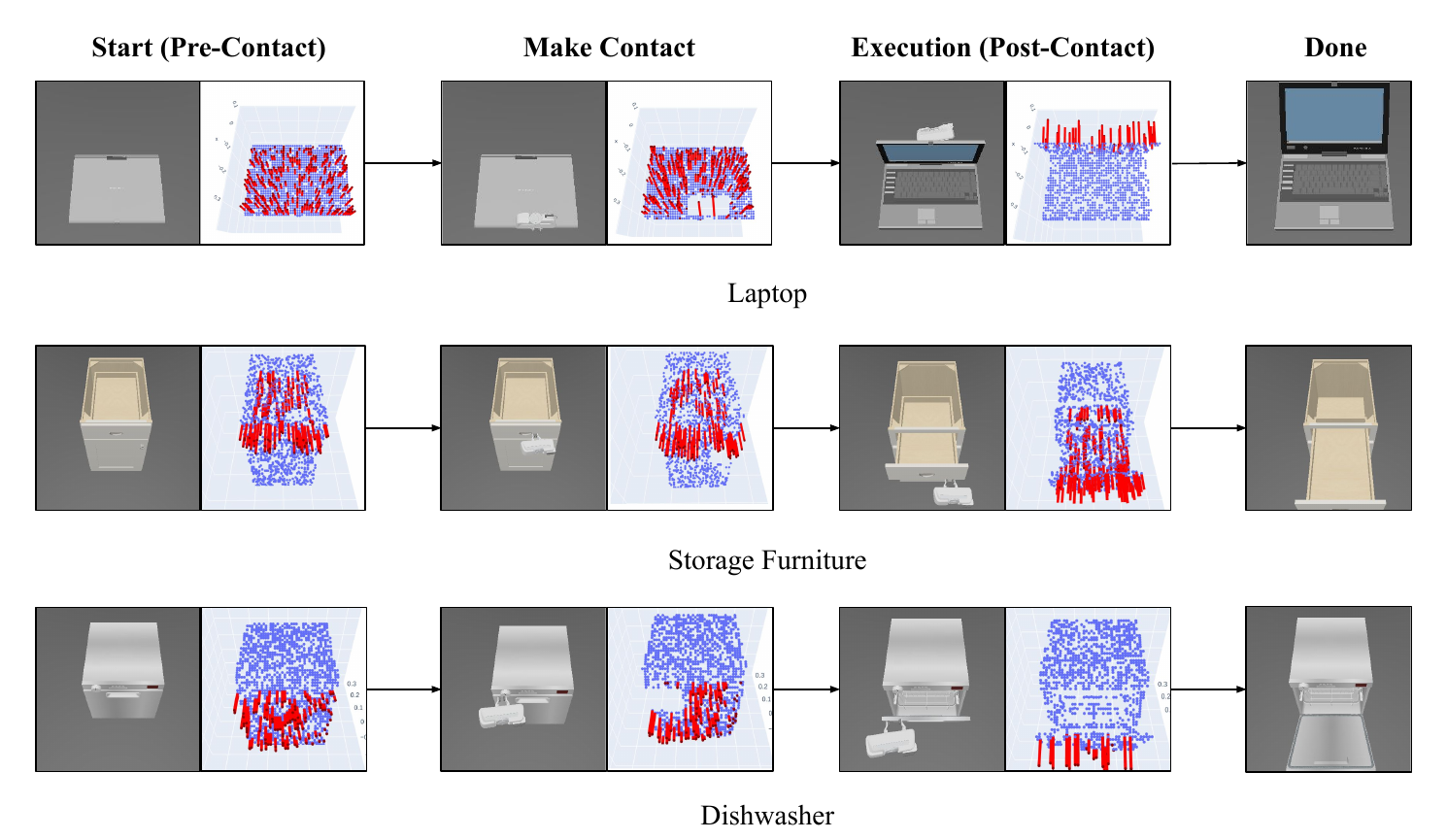}
\caption{Simulated rollout examples}
\label{tab:sim-pipeline}
\end{figure*}

\section{Real-World Dataset}
As shown in Section IV-B, we use 14 different objects in real world experiments. The objects labeled 1-14 in Fig. 5 are described here in Table \ref{tab:labels}.
\begin{table}[htb]
\centering
\begin{tabular}{|c|c|c|c|}
\hline
Label&ID&Category&Type\\
\hline
1&  \texttt{chest\_1} & Box & Revolute \\
2&    \texttt{teapot\_1} & Kettle & Prismatic\\
3&    \texttt{toilet\_1} & Toilet & Revolute\\
4&    \texttt{fridge\_1} & Refrigerator & Revolute\\
5&    \texttt{oven\_1} & Oven & Revolute\\
6&    \texttt{drawer\_1} & Storage & Prismatic\\
7&    \texttt{safe\_1} & Safe &  Revolute\\
8&    \texttt{microwave\_1} & Microwave & Revolute \\
9&    \texttt{minifridge\_1} & Refrigerator & Revolute\\
10&    \texttt{jar\_1} & Kitchen Pot & Prismatic\\
11&    \texttt{jar\_2} & Kitchen Pot &Prismatic\\
12&    \texttt{trash\_1} & Trash Can &Revolute\\
13&    \texttt{laptop\_1} & Laptop &  Revolute\\
14&    \texttt{box\_1} & Box &Revolute\\
 \hline
\end{tabular}
\caption{Labels and their corresponding objects and the objects' articulation types shown in Fig. 5 of the paper. Note that \texttt{jar\_1} and \texttt{jar\_2} are not technically kitchen pots but they do have lids similar to kitchen pots in functionality and have identical ariculation parameters.}
\label{tab:labels}
\end{table}

\begin{figure*}[ht]
    \centering
    \begin{tabular}{cccc}
        \subfloat[box\_1]{\includegraphics[width = .20\linewidth]{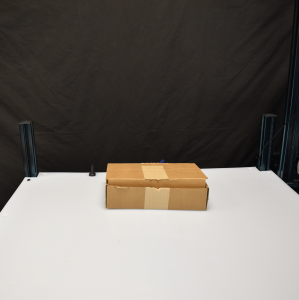}} &
        \subfloat[chest\_1]{\includegraphics[width = .20\linewidth]{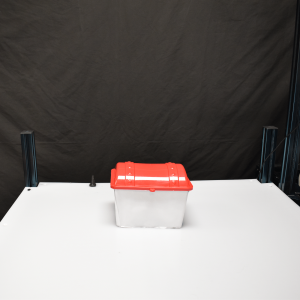}} &
        \subfloat[drawer\_1]{\includegraphics[width = .20\linewidth]{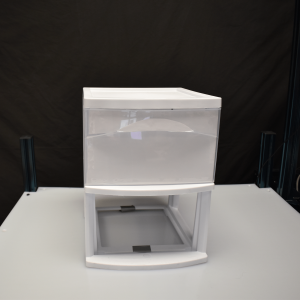}} \\[0pt]
        \subfloat[fridge\_1]{\includegraphics[width = .20\linewidth]{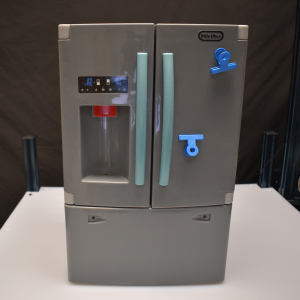}} &
        \subfloat[jar\_1]{\includegraphics[width = .20\linewidth]{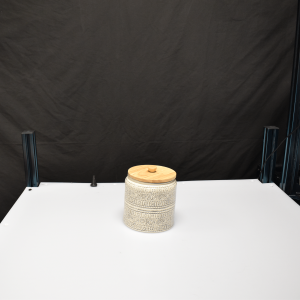}} &
        \subfloat[jar\_2]{\includegraphics[width = .20\linewidth]{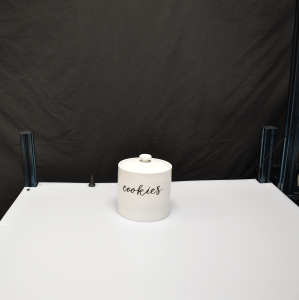}} \\[0pt]
        \subfloat[laptop\_1]{\includegraphics[width = .20\linewidth]{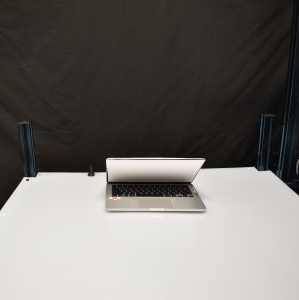}} &
        \subfloat[microwave\_1]{\includegraphics[width = .20\linewidth]{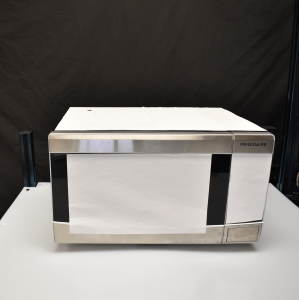}} &
        \subfloat[minifridge\_1]{\includegraphics[width = .20\linewidth]{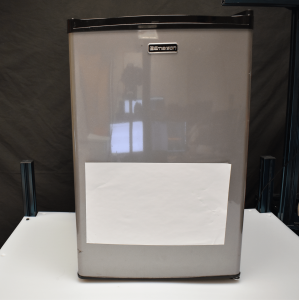}} \\[0pt]
        \subfloat[oven\_1]{\includegraphics[width = .20\linewidth]{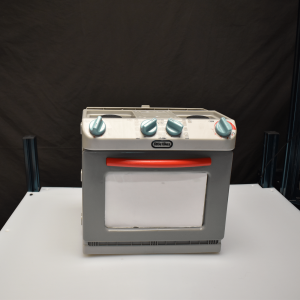}} &
        \subfloat[safe\_1]{\includegraphics[width = .20\linewidth]{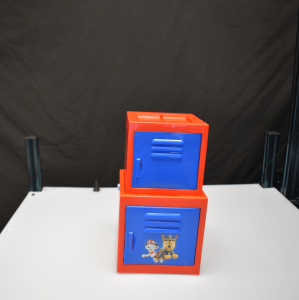}} &
        \subfloat[teapot\_1]{\includegraphics[width = .20\linewidth]{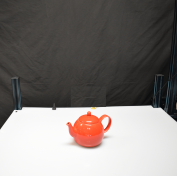}} \\[0pt]
        \subfloat[toilet\_1]{\includegraphics[width = .20\linewidth]{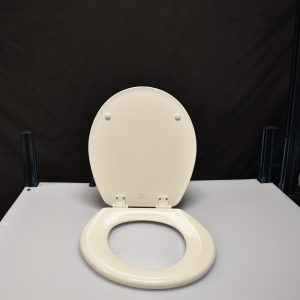}} &
        \subfloat[trashcan\_1]{\includegraphics[width = .20\linewidth]{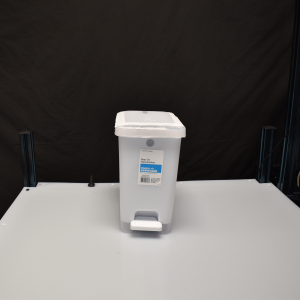}} &
        \end{tabular}
        \caption{Objects in the dataset for real world experiments}
    \label{fig:allobjs}
\end{figure*}

In Fig. \ref{fig:allobjs}, we show the 14 objects individually for more clarity. 

\section{Results in the UMPNet Environment}

\hl{We perform a direct evaluation of FlowBot3D in the UMPNet simulation environment, which was only made public after this manuscript was accepted for publication. There are several major differences between our main simulation environment and the UMPNet evaluation environment:}

\begin{itemize}
    \item The UMPNet environment uses the PyBullet physics simulator, whereas we use the SAPIEN environemnt (backed by PhysX).
    \item The UMPNet environment disables collisions between the gripper geometry and the rest of the object (except for the part where contact is made). We leave full contact enabled.
    \item The UMPNet environment has a hard contact constraint between the object and the gripper, whereas our contact is softer, acting more like a spring.
\end{itemize}

\begin{table*}[ht]
\renewcommand{\arraystretch}{1.2}
\resizebox{\textwidth}{!}{
\setlength\tabcolsep{.2em}
\begin{tabular}{|r|c|ccccccccccc||c|cccccccccc|}
 \hline
 \multicolumn{13}{|c||}{\textbf{Novel Instances in Train Categories}} 
 &
 \multicolumn{11}{c|}{\textbf{Test Categories}}
 \\
\hline
\rule{0pt}{2.5em}
 & \textbf{\underline{AVG.}} &\clipart{stapler} &\clipart{trash}&\clipart{storage}&\clipart{window}&\clipart{toilet}&\clipart{laptop}&\clipart{kettle}&\clipart{switch}&\clipart{fridge}&\clipart{chair}&\clipart{microwave}&\textbf{\underline{AVG.}}&\clipart{bucket}&\clipart{safe}&\clipart{phone}&\clipart{pot}&\clipart{box}&\clipart{table}&\clipart{dish}&\clipart{oven}&\clipart{washer}&\clipart{door}\\
 \hline
 \textbf{Baselines} & \multicolumn{23}{c|}{} \\
 \hline 
 UMPNet&0.18&\textbf{0.18}&\textbf{0.17}&0.32&0.32&0.05&0.06&\textbf{0.12}&\textbf{0.24}&0.23&0.18&0.08&\textbf{0.15}&0.23&\textbf{0.14}&0.04&\textbf{0.00}&0.25&0.27&0.09&0.21&\textbf{0.13}&\textbf{0.19}\\
 \hline

 \textbf{Ours} & \multicolumn{23}{c|}{} \\
 \hline 
  FlowBot3D in UMPNet&\textbf{0.17}&0.42&0.22&\textbf{0.16}&\textbf{0.17}&\textbf{0.03}&\textbf{0.00}&0.20&0.51&\textbf{0.07}&\textbf{0.00}&\textbf{0.08}&0.21&\textbf{0.17}&0.29&\textbf{0.00}&0.06&\textbf{0.21}&\textbf{0.10}&\textbf{0.06}&\textbf{0.16}&0.29&0.73\\
  \hline
\end{tabular}
}
\smallskip

\caption{Normalized Distance Metric Results: Normalized distances evaluated in the official UMPNet environment to the target articulation joint angle after a full rollout across different methods. The lower the better.}
\label{tab:baselines-dist}
\end{table*}
\begin{table*}[h]
\renewcommand{\arraystretch}{1.2}
\resizebox{\textwidth}{!}{
\setlength\tabcolsep{.2em}
\begin{tabular}{|r|c|ccccccccccc||c|cccccccccc|}
 \hline
 \multicolumn{13}{|c||}{\textbf{Novel Instances in Train Categories}} 
 &
 \multicolumn{11}{c|}{\textbf{Test Categories}}
 \\
\hline
\rule{0pt}{2.5em}
 & \textbf{\underline{AVG.}} &\clipart{stapler} &\clipart{trash}&\clipart{storage}&\clipart{window}&\clipart{toilet}&\clipart{laptop}&\clipart{kettle}&\clipart{switch}&\clipart{fridge}&\clipart{chair}&\clipart{microwave}&\textbf{\underline{AVG.}}&\clipart{bucket}&\clipart{safe}&\clipart{phone}&\clipart{pot}&\clipart{box}&\clipart{table}&\clipart{dish}&\clipart{oven}&\clipart{washer}&\clipart{door}\\
 \hline
 \textbf{Baselines} & \multicolumn{23}{c|}{} \\
 \hline 
 UMPNet&0.73&\textbf{0.73}&0.71&0.60&0.49&0.89&0.90&0.79&\textbf{0.60}&0.64&0.78&0.86&\textbf{0.75}&0.55&\textbf{0.80}&0.89&\textbf{1.00}&0.66&0.64&\textbf{0.77}&0.64&\textbf{0.75}&\textbf{0.76}\\
 \hline

 \textbf{Ours} & \multicolumn{23}{c|}{} \\
 \hline 
 FlowBot3D in UMPNet&\textbf{0.81}&0.53&\textbf{0.74}&\textbf{0.81}&\textbf{0.82}&\textbf{0.96}&\textbf{0.99}&\textbf{0.79}&0.44&\textbf{0.90}&\textbf{1.00}&\textbf{0.89}&0.70&\textbf{0.69}&0.63&\textbf{1.00}&0.94&\textbf{0.67}&\textbf{0.89}&0.75&\textbf{0.66}&0.69&0.14\\
  \hline
\end{tabular}
}
\smallskip

\caption{Success Rate Metric Results: Fraction of success trials (normalized distance less than 0.1) of different objects' categories after a full rollout across different methods evaluated in the official UMPNet environment. The higher the better. }
\vspace*{-10pt}
\label{tab:baselines-sr}
\end{table*}

We use the UMPNet evaluation script without modification, with the exception that the chosen action is selected based on FlowBot3D instead. In Tables \ref{tab:baselines-dist} and \ref{tab:baselines-sr}, we present the results for the following methods: 

\begin{itemize}
    \item\textbf{UMPNet}: We run a pre-trained UMPNet model with the official UMPNet code following the exact same evaluation procedure listed in [33]. The numbers here are consistent with those in the UMPNet paper.
 
    \item \textbf{FlowBot3D in UMPNet Environment}: FlowBot3D trained and evaluated with the camera parameters and objects' placement randomization from UMPNet's PyBullet environment. Note that in test time, UMPNet takes as input a goal of the articulated object in its fully closed or fully open state, so we use the ground-truth goal to decide if we need to invert the output 3DAF directions (i.e. if the ground-truth goal is a fully closed state, we invert the output direction).
\end{itemize}

Overall, the two methods perform similarly on the task. However, while the ArtFlowNet was retrained on point clouds generated in PyBullet, the performance was not significantly tuned on the different task distribution in the UMPNet dataset.

\section{Full Trials Results}
In Table \ref{tab:supp-rw-flow} and \ref{tab:supp-rw-daggero}, we show the full trials results, which contains the metrics averaged over all 5 trials for each object.

\clearpage

\clearpage

\begin{table*}
\centering
\begin{tabular}{|c|cccccc|}
 \hline
 Object ID & Object Category & Success/Total & Success \% & Contact Success/Total & Distance & Motion-Only Success/Total \\
 \hline
    \texttt{chest\_1} & Box & 3/5 & 60\% & 4/5 & 0.22 & 3/4 \\
    \texttt{teapot\_1} & Kettle & 5/5 & 100\% & 5/5 & 0.00 & 5/5 \\
    \texttt{toilet\_1} & Toilet & 4/5 & 80\% & 5/5 & 0.02 & 4/5 \\
    \texttt{fridge\_1} & Refrigerator & 3/5 & 60\% & 3/5 & 0.11 & 5/5 \\
    \texttt{oven\_1} & Oven & 0/5 & 0\% & 5/5 & 1.00 & 0/5 \\
    \texttt{drawer\_1} & Storage & 3/5 & 60\% & 3/5 & 0.40 & 3/3 \\
    \texttt{safe\_1} & Safe & 1/5 & 20\% & 2/5 & 0.73 & 1/2 \\
    \texttt{microwave\_1} & Microwave & 3/5 & 60\% & 5/5 & 0.11 & 3/5 \\
    \texttt{minifridge\_1} & Refrigerator & 2/5 & 40\% & 5/5 & 0.155 & 2/5 \\
    \texttt{jar\_1} & Kitchen Pot & 5/5 & 100\% & 5/5 & 0.00 & 5/5 \\
    \texttt{jar\_2} & Kitchen Pot & 5/5 & 100\% & 5/5 & 0.00 & 5/5 \\
    \texttt{trash\_1} & Trash Can & 5/5 & 100\% & 5/5 & 0.02 & 5/5 \\
    \texttt{laptop\_1} & Laptop & 4/5 & 100\% & 5/5 & 0.07 & 4/5 \\
    \texttt{box\_1} & Box & 2/5 & 40\% & 5/5 & 0.28 & 2/5 \\
    
 \hline
 \hline
 \textbf{SUMMARY} & \textbf{-} & \textbf{45/70} & \textbf{64.3\%} & \textbf{64/70} & \textbf{0.22} & \textbf{45/64} \\
 \hline
\end{tabular}
\smallskip
\caption{Real-World Trials for FlowNet}
\label{tab:supp-rw-flow}
\end{table*}
\begin{table*}
\centering
\begin{tabular}{|c|cccccc|}
 \hline
 Object ID & Object Category & Success/Total & Success \% & Contact Success/Total & Distance & Motion-Only Success/Total \\
 \hline
    \texttt{chest\_1} & Box & 1/5 & 20\% & 5/5 & 0.80 & 1/5 \\
    \texttt{teapot\_1} & Kettle & 2/5 & 40\% & 5/5 & 0.60 & 2/5 \\
    \texttt{toilet\_1} & Toilet & 0/5 & 0\% & 5/5 & 0.78 & 0/5 \\
    \texttt{fridge\_1} & Refrigerator & 0/5 & 0\% & 5/5 & 1.00 & 0/5 \\
    \texttt{oven\_1} & Oven & 0/5 & 0\% & 5/5 & 1.00 & 0/5 \\
    \texttt{drawer\_1} & Storage & 1/5 & 20\% & 5/5 & 0.72 & 1/5 \\
    \texttt{safe\_1} & Safe & 1/5 & 20\% & 3/5 & 0.70 & 1/3 \\
    \texttt{microwave\_1} & Microwave & 0/5 & 0\% & 5/5 & 1.00 & 0/5 \\
    \texttt{minifridge\_1} & Refrigerator & 0/5 & 0\% & 5/5 & 1.00 & 0/5 \\
    \texttt{jar\_1} & Kitchen Pot & 3/5 & 60\% & 5/5 & 0.40 & 3/5 \\
    \texttt{jar\_2} & Kitchen Pot & 1/5 & 20\% & 5/5 & 0.80 & 1/5 \\
    \texttt{trash\_1} & Trash Can & 0/5 & 0\% & 5/5 & 1.00 & 0/5 \\
    \texttt{laptop\_1} & Laptop & 1/5 & 20\% & 5/5 & 0.81 & 1/5 \\
    \texttt{box\_1} & Box & 1/5 & 20\% & 5/5 & 0.80 & 1/5 \\
    
 \hline
 \hline
 \textbf{SUMMARY} & \textbf{-} & \textbf{10/70} & \textbf{14.3\%} & \textbf{68/70} & \textbf{0.73} & \textbf{10/68} \\
 \hline
\end{tabular}
\smallskip
\caption{Real-World Trials for DAgger Oracle}
\label{tab:supp-rw-daggero}
\end{table*}
\clearpage



\end{document}